# Learning Bayesian networks that enable full propagation of evidence

Anthony C. Constantinou[1,2]

[1]Bayesian Artificial Intelligence research lab, Risk and Information Management (RIM) Research Group, School of EECS, Queen Mary University of London (QMUL), London, UK, E1 4NS.
[2] The Alan Turing Institute, UK.

E-mail: a.constantinou@qmul.ac.uk.

This research was supported by the ERSRC Fellowship project EP/S001646/1 on Bayesian Artificial Intelligence for Decision Making under Uncertainty, and by The Alan Turing Institute in the UK under the EPSRC grant EP/N510129/1.

**ABSTRACT** This paper builds on recent developments in Bayesian network (BN) structure learning under the controversial assumption that the input variables are dependent. This assumption can be viewed as a learning constraint geared towards cases where the input variables are known or assumed to be dependent. It addresses the problem of learning multiple disjoint subgraphs that do not enable full propagation of evidence. This problem is highly prevalent in cases where the sample size of the input data is low with respect to the dimensionality of the model, which is often the case when working with real data. The paper presents a novel hybrid structure learning algorithm, called SaiyanH, that addresses this issue. The results show that this constraint helps the algorithm to estimate the number of true edges with higher accuracy compared to the state-of-the-art. Out of the 13 algorithms investigated, the results rank SaiyanH 4[th] in reconstructing the true DAG, with accuracy scores lower by 8.1% (F1), 10.2% (BSF), and 19.5% (SHD) compared to the top ranked algorithm, and higher by 75.5% (F1), 118% (BSF), and 4.3% (SHD) compared to the bottom ranked algorithm. Overall, the results suggest that the proposed algorithm discovers satisfactorily accurate connected DAGs in cases where other algorithms produce multiple disjoint subgraphs that often underfit the true graph.

**INDEX TERMS** causal discovery, conditional independence, directed acyclic graphs, probabilistic graphical models, structure learning.

## I. INTRODUCTION

A Bayesian Network (BN) is a type of a probabilistic graphical model introduced by Pearl [1] [2]. If we assume that the arcs between nodes represent causation, then the BN is viewed as a Causal Bayesian Network (CBN). However, if we assume that the edges between nodes represent some dependency that is not necessarily causal, then such a BN is viewed as a dependence graph. A CBN can only be represented by a unique Directed Acyclic Graph (DAG), whereas a BN that is not viewed as a causal model can be also be represented by a Completed Partial Directed Acyclic Graph (CPDAG). A CPDAG incorporates both directed and undirected edges and represents a set of Markov equivalent DAGs that entail the same independence relations over the observed variables.

BNs have emerged as one of the most successful approaches for reasoning under uncertainty. This is partly because they enable decision makers to reason with transparent causal assumptions that offer solutions that go beyond prediction. For example, a CBN enables decision makers to reason about intervention and counterfactuals. On this basis, the focus of this paper is on the reconstruction of the true causal DAG, as opposed to the reconstruction of a graph that forms part of the equivalence class of the true DAG (i.e., a CPDAG).

Constructing a BN involves determining the graphical structure of the network and parameterising its conditional distributions. The problem of structure learning is considerably more challenging than that of parameter learning. This is because searching for the optimal graph represents an NP-Hard problem where some instances are much harder than others [3]. Structure learning algorithms generally fall under two learning classes. Firstly, the score-based methods represent a traditional machine learning approach where graphs are explored and scored in terms of how well the fitting distributions agree with the empirical distributions. The graph that maximises the scoring function is returned as the preferred graph. On the other hand, constraint-based learning is based on a series of conditional independence tests that determine the removal and the orientation of some edges. Hybrid algorithms are often viewed as a third learning classe that adopts features from both score-based and constraint-based learning.

The automated construction of causal structures has the potential to offer significant benefits to every research field concerned with causal inference and actions for intervention. However, automated causal discovery is hindered by difficulties that have significantly limited its impact. These difficulties go beyond the problem of NP-hardness that is generally addressed by algorithms that prune the search space of possible graphs and effectively minimise the loss in accuracy and maximise the gain in speed.

Importantly, there are conflicting claims in the literature about what can be recovered from observational data. Some argue for a causal graph and others for a dependence graph [4] [5] [6] [7] [8]. The underlying assumption of the learned graph influences the evaluation process that determines the effectiveness of these algorithms. While cross-validation serves as an excellent evaluator for a predictive model in other machine learning fields, it underdetermines the accuracy of causal inference. As a result, there is no consensus on an evaluation approach that best determines the effectiveness of a BN structure learning algorithm. Each publication makes an empirical or a theoretical case for the algorithm presented in that publication [9]. Likewise, each structure learning algorithm is based on a set of assumptions, such as complete data and causal sufficiency, and tends to be evaluated with synthetic data that conforms to those assumptions, however unrealistic these assumptions may be in the real world [10]. Because of this, it is widely accepted that synthetic performance overestimates real performance. These unresolved issues continue to invite different forms of domain knowledge to be incorporated into the structure learning process [11] [12] [13] [14] [15]. The learning constraint proposed in this paper can be viewed as one more such knowledge-based constraint.

The rest of the paper is structured as follows: Section 2 describes the algorithm, Section 3 describes and discusses the evaluation process, Section 4 presents and discusses the results, and Section 5 provides the concluding remarks along with possible directions for future research.

## II. THE ALGORITHM

The algorithm addresses the problem of learning multiple disjoint subgraphs that do not enable full propagation of evidence. This is achieved by performing structure learning under the assumption that the input variables are dependent.

The learning process of the algorithm consists of three phases. The first phase starts by producing an initial best guess undirected graph that is entirely based on pairwise associational scores. Constraint-based learning is then used in conjunction with other rules to orientate edges in phase 2. The third and final phase involves score-based learning that modifies the graph produced at phase 2 towards the path that maximises a scoring function. The subsections that follow describe these three phases in turn, as well as the computational complexity of the algorithm.

### A. Phase 1: Associational learning

The first phase is based on two novel approaches inherited from an early experimental version of this algorithm [16]. They involve a) the associational score Mean/Max/MeanMax Marginal Discrepancy (MMD), and b) an undirected graph called the Extended Maximum Spanning Graph (EMSG). The output of phase 1 is the EMSG and serves as the starting graph of phase 2.

The MMD score represents the discrepancy in marginal probabilities between prior and posterior distributions. Contrary to other traditional measures such as mutual information (MI), the MMD score offers linear examination of the marginal and conditional independencies[1]. The MMD score ranges from 0 to 1, where a higher score indicates a stronger dependency. For edge A↔B, the score $MMD(A \leftrightarrow B)$ is the average of scores $MMD_{MN}(A \leftrightarrow B)$ and $MMD_{MX}(A \leftrightarrow B)$, where $MN$ and $MX$ are mean and max marginal discrepancies. Specifically,

$$MMD(A \leftrightarrow B) = \sum_{\leftrightarrow} \sum_{m} MMD_m(A \leftrightarrow B)w$$

where $\leftrightarrow$ represents the iterations over $\leftarrow$ and $\rightarrow$, $m$ represents the iterations over $MN$ and $MX$, and $w$ is the normalising constant 0.25 for the scores accumulated over the following four iterations:

$$MMD_{MN}(A \rightarrow B) = \left( \sum_{j}^{s_A} \left[ \left( \sum_{i}^{s_B} |P(B_i) - P(B_i|A_j)| \right) / s_B \right] \right) / s_A$$

$$MMD_{MN}(A \leftarrow B) = \left( \sum_{i}^{s_B} \left[ \left( \sum_{j}^{s_A} |P(A_j) - P(A_j|B_i)| \right) / s_A \right] \right) / s_B$$

$$MMD_{MX}(A \rightarrow B) = \left( \sum_{j}^{s_A} \max_{i} |P(B_i) - P(B_i|A_j)| \right) / s_A$$

---

[1]This does not imply that the MMD score is superior to the MI score or other non-linear associational measure.

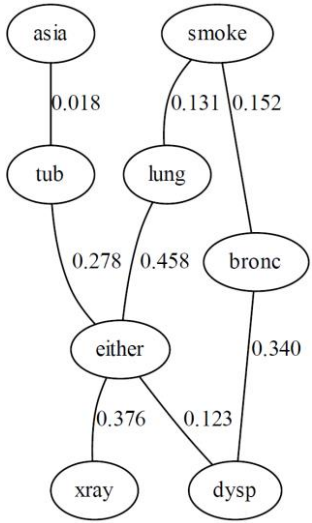

**FIGURE 1.** The EMSG based on the Asia BN example, with the MMD scores produced at the end of phase 1.

$$MMD_{MX}(A \leftarrow B) = \left( \sum_{i}^{s_B} \max_j |P(A_j) - P(A_j|B_i)| \right) / s_B$$

for each state $j$ in $A$ and state $i$ in $B$, and over the $S_A$ states in $A$ and $S_B$ states in $B$.

The EMSG is determined by the MMD scores and can be viewed as an extended version of the maximum spanning tree [17]. This is because EMSG preserves multiple connecting paths from one node to another, unlike the maximum spanning tree which preserves the single and most likely connecting path between nodes. The intention here is to start with a graph that is more dense, in terms of the number of edges, compared to the corresponding maximum spanning tree.

Starting from a complete graph, the EMSG is produced by removing edges between two nodes $A$ and $B$ if and only if $A$ and $B$ share neighbour $C$ where

$$MMD(A \leftrightarrow C) > MMD(A \leftrightarrow B) < MMD(B \leftrightarrow C)$$

The order in which the edges are assessed for removal is from lowest to highest MMD score. Figure 1 presents the EMSG produced for the classic Asia BN, along with the MMD scores assigned to each of the edges. In this example, the EMSG matches the skeleton of the true Asia graph.

### B. Phase 2: Constraint-based learning

In the second phase, SaiyanH performs conditional independence tests across all pairs of nodes conditional on the remaining nodes in sets of triples, and classifies each triple into either conditional dependence, independence or insignificance. Assuming independence tests between $A$ and $B$ conditional on $C$, the following rules apply for classification:

1. **Conditional dependence:** if $MMD(A \leftrightarrow B)|C$ is both greater than 0.05 and 50% higher than $MMD(A \leftrightarrow B)$.

2. **Conditional independence:** if $MMD(A \leftrightarrow B)|C$ is both lower than 0.05 and 50% lower than $MMD(A \leftrightarrow B)$.

These thresholds represent the hyperparameter defaults adopted by other algorithms that employ similar processes to investigate independence. Specifically, the dependency threshold of 0.05 corresponds to the same cut-off threshold of the unoptimised parameter *alpha* used in other constraint-based algorithms [18] [19]. The additional threshold of 50% represents a new rule used for conditional independence tests that lead to more conservative classifications of conditional independence. This rule produces a higher number of conditional independence tests classified as 'conditional insignificance' and produces fewer, although more certain, conditional dependence and independence classifications of triples. The unoptimised rate of 50% represents a hypermarameter default that is analogous to the default threshold of 0.5 in RFCI-BSC used to determine whether the constraints are dependent [20], and to the default threshold of 0.5 in CCHM used to analyse causal effects [21]. The classifications from constraint-based learning partly determine the orientation of the edges in EMSG during phase 2, and are also used to prune the search space of graphs explored in phase 3 (refer to subsection II.C).

The order in which the edges in EMSG are assessed for orientation is determined by node ordering, where nodes are ordered by the total MMD score they share with their neighbours. For example, the starting node in the EMSG graph of Fig 1 would be the node 'either' because it shares a total score of 1.235 with its neighbouring nodes, and which is the highest total score over all the nodes in the network. Once a node is selected, the edges of that node are evaluated in the order they appear in the data. If an orientation leads to a cyclic graph, the orientation of that edge is immediately reversed under the assumption that preceding orientations override proceeding results.

The orientation of the edges in EMSG is based on a set of criteria. The conditional independence classifications serve as the first criterion. Specifically, if the conditional dependence and independence classifications support an orientation, then the edge under assessment is orientated. Otherwise, the edge under assessment remains undirected and the algorithm proceeds to the next edge. Edges that remain undirected are re-assessed, in the same order, with the second criterion which is the BIC score (refer to subsection II.C). However, the BIC score is score-equivalent and there is no formal guarantee that all edges will be recovered by this second criterion. Edges that continue to be undirected are then re-assessed with a third criterion, the *do*-calculus [22], which is used to maximise the number of nodes influenced by intervention. For example, in assessing the undirected edge $A - B$, if $do(a)$ given $A \rightarrow B$ influences a higher number of nodes (i.e., children and descendants) than $do(b)$ given $A \leftarrow B$, then the algorithm will orientate the edge $A - B$ as $A \rightarrow B$. If some edges continue to remain undirected at the end of this process, the undirected edges are re-assessed with the second and third criteria.

### C. Phase 3: Score-based learning

The output of phase 2 serves as the starting graph for score-based learning in phase 3. SaiyanH uses the BIC to score the DAGs being explored. The BIC is a model selection function that balances model fitting with model dimensionality given the data. Formally,

$$BIC = LL(G|D) - \left(\frac{log_2 N}{2}\right)p$$

for graph $G$ given data $D$, $LL$ is the log-likelihood, $N$ is the sample size of $D$, and $p$ is the number of free parameters (also known as independent parameters) in $G$. Specifically, the number of free parameters $p$, which is as a measure of model dimensionality, is:

$$p = \sum_i^{|V|}(r_i - 1)\prod_j^{|\pi_{v_i}|} q_j$$

where paper $V$ is a set of the variables $v_i$ in graph $G$, $|V|$ is the size of set $V$, $r_i$ is the number of states of $v_i$, $\pi_{v_i}$ is the parent set of $v_i$, $|\pi_{v_i}|$ is the size of set $\pi_{v_i}$, and $q_j$ is the number of states of $v_j$ in parent set $\pi_{v_i}$.

The search starts with Hill-Climbing (HC) that explores neighbouring graphs $G'$ in which an edge is reversed, removed, or added. Whenever $G'$ has BIC greater than $G$, $G$ is replaced with $G'$. This process continues until no neighbour $G'$ increases the BIC score. When HC completes search, attempts are made to escape possible local maxima using Tabu search. This is achieved by examining if a neighbour $G'$ that minimally decreases BIC has a neighbour $G''$ that improves the BIC score of $G$, in which case $G$ is replaced with $G''$ and Tabu search restarts at the new $G$. When all $G''$ for a particular $G'$ are explored without further improvement in the BIC score of $G$, the search proceeds to the next best $G'$ that minimally decreases BIC. Tabu search completes when all $G'$ are explored, or when the number of escape attempts $G'$ reaches $|V|(|V| - 1)$.

The search space of possible graphs is restricted to graphs that are acyclic and to graphs that do not consist of multiple disjoint subgraphs. Moreover, as briefly discussed in subsection II.B, the search space of arc additions is pruned by means of marginal and conditional independence. Specifically, arcs with MMD $< 0.05$ and arcs that violate any

---

**Algorithm 1:** SaiyanH pseudocode

**Input**: dataset $D$, a fully connected graph $G$, score function BIC($G, D$)
**Output**: graph $G$

// **Phase 1**
1: **for** each pair of variables $v_i, v_j \in D$ **do**
2:    add $v_i \leftrightarrow v_j$ with score MMD$(v_i \leftrightarrow v_j)$ to list $M$ in ascending order
3: **end for**
4: **for** each $v_i \leftrightarrow v_j \in M$ **do**
5:    **if** MMD$(v_i \leftrightarrow v_k)_i >$ MMD$(v_i \leftrightarrow v_j)_i <$ MMD$(v_j \leftrightarrow v_k)_i$ **then**
6:      remove edge $v_i \leftrightarrow v_j$ in $G$
7:    **end if**
8: **end for**

// **Phase 2**
9: **for** each pair of variables $v_i, v_j$ conditional on $v_k \in D$ **do**
10:    **if** $0.05 <$ MMD$(v_i \leftrightarrow v_j|v_k) >$ MMD$(v_i \leftrightarrow v_j) \times 1.5$ **then**
11:      add $v_i \leftrightarrow v_j$ with score MMD$(v_i \leftrightarrow v_j|v_k)$ to list $CD$
12:    **else if** $0.05 >$ MMD$(v_i \leftrightarrow v_j|v_k) <$ MMD$(v_i \leftrightarrow v_j) \times 0.5$ **then**
13:      add $v_i \leftrightarrow v_j$ with score MMD$(v_i \leftrightarrow v_j|v_k)$ to list $CI$
14:    **end if**
15: **end for**
16: **for** each edge $v_i \leftrightarrow v_j \in G$ **do** (in ascending order MMD)
17:    **if** $CD$ and $CI$ support an orientation for edge $v_i \leftrightarrow v_j$ in $G$ **then**
18:      orientate edge $v_i \leftrightarrow v_j$ (reverse edge if acyclicity is violated)
19:      **if** orientation of edge $v_i \leftrightarrow v_j$ violates acyclicity in $G$ **then**
20:        reverse the orientation of edge $v_i \leftrightarrow v_j$
21:      **end if**
22:    **end if**
23: **end for**
24: **while** an undirected edge $v_i \leftrightarrow v_j \in G$ **do**
25:    **for** each undirected edge $v_i \leftrightarrow v_j \in G$ **do** (in ascending order MMD)
26:      **if** an orientation of edge $v_i \leftrightarrow v_j$ maximises BIC($G, D$) **then**
27:        orientate edge $v_i \leftrightarrow v_j$ (reverse edge if acyclicity is violated)
28:      **end if**
29:    **end for**
30:    **for** each undirected edge $v_i \leftrightarrow v_j \in G$ **do** (in ascending order MMD)
31:      **if** an orientation of edge $v_i \leftrightarrow v_j$ maximises $do(v_i \leftrightarrow v_j, G)$ **then**
32:        orientate edge $v_i \leftrightarrow v_j$ (reverse edge if acyclicity is violated)
33:      **end if**
34:    **end for**
35: **end while**

// **Phase 3**
36: **while** Hill-Climbing finds BIC($G', D$) > BIC($G, D$) **and**
     marginal and conditional independencies are not violated **and**
     $G'$ is a valid DAG **do**
37:    $G = G'$
38: **end while**
39: **while** Tabu finds BIC($G'', D$) > BIC($G, D$) **and**
     marginal and conditional independencies are not violated **and**
     single-depth Tabu escapes $e$ have not been explored **and**
     $e < |V|(|V| - 1)$ **do**
40:    $G = G''$
41: **end while**

---

conditional independence classification, as defined in subsection II.B, are pruned and hence not explored. Algorithm 1 presents the pseudocode of SaiyanH.

### D. Computational complexity

The complexity of local learning and constraint-based learning is generally determined by the number of local and conditional associational tests executed by an algorithm [4]. Given a variable set $V$, the complexity of local learning $O_L$ in phase 1 of SaiyanH is:

$$O_L = \left(\frac{|V|(|V|-1)}{2}\right)$$

whereas the complexity of constraint-based learning $O_C$ during phase 2 is [16]:

$$O_C = \left(\frac{|V|(|V|-1)(|V|-2)}{2}\right)$$

On the other hand, the score-based learning (i.e., phase 3) is based on Tabu search which is a metaheuristic. In BN structure learning, a metaheuristic such as Tabu search depends on the number of local maxima that surround the initial best guess graph (e.g., the output of phase 2), and these can vary greatly given the data. As a result, the theoretical complexity of metaheuristics cannot be expressed accurately with traditional complexity notions. According to the timing results shown later in Table 4, score-based learning has complexity $O_S$ which can be lower or higher than $O_L$ and $O_C$ depending on the sample size of the input data. Empirical results show that $O_L < \sim O_S > O_C$ when sample size of the input data is low relative to those considered in this paper, $O_L < \sim O_S < O_C$ when the sample size is moderate, and $O_L > \sim O_S < O_C$ when the sample size is high.

## III. EVALUATION

### A. Scoring metrics

The evaluation of BN structure learning algorithms is generally based on metrics that assess the relevance of the learned graph with respect to the ground truth graph. Less often, the evaluation may be based on measures which determine how well the learned distributions fit the data. However, fitting scores are generally score-equivalent and produce the same score for Markov equivalent DAGs. Because the scope of this paper focuses on the reconstruction of the true DAG, the scoring criteria considered are fully oriented towards graphical discovery.

Three different scoring metrics are considered that make varied use of the confusion matrix parameters. The differences between these three metrics can often highlight advantages and disadvantages of an algorithm that would otherwise remain unknown. Since no metric is perfect, using multiple metrics provides a fairer comparison between algorithms.

The three metrics use varying combinations of the following parameters [23]:

- *True Positives* (TP): The number of edges discovered in the learned graph that exist in the true graph.

- *True Negatives* (TN): The number of direct independencies discovered in the learned graph that exist in the true graph.

- *False Positives* (FP): The number of edges discovered in the learned graph that do not exist in the true graph.

- *False Negatives* (FN): The number of direct independencies discovered in the learned graph that do not exist in the true graph.

Moreover, edges in the learned graph that fail to produce the correct orientation, including undirected and bi-directed edges produced by some of the other algorithms, receive 50% reward relative to the reward allocated to the edge with the correct orientation. The first metric, the F1 score, is based on both the Recall ($Re$) and Precision ($Pr$) which are the two standard metrics used in this research field. Specifically,

$$Pr = \frac{TP}{TP+FP} \qquad Re = \frac{TP}{TP+FN}$$

The Recall and Precision scores are, however, misleading when reported independently. The F1 score, on the other hand, offers the harmonic mean between the two:

$$F1 = 2\frac{Pr.Re}{Pr+Re}$$

where F1 ranges from 0 to 1 and a higher score indicates a more accurate graph.

The second metric, called the Structural Hamming Distance (SHD) [24], is another well-established metric in this field of research and represents the number of steps required to transform the learned graph into the ground truth graph. Specifically,

$$SHD = FP + FN$$

where a score of 0 indicates a perfect match between the learned graph and the true graph.

The third metric, called the Balanced Scoring Function (BSF), is a recent metric [23] that considers all the four confusion matrix parameters and returns a fully balanced score. The score ranges from -1 to 1, where -1 corresponds to the worst possible graph, 1 to the graph that matches the true graph, and 0 to an empty or a fully connected baseline graph. Specifically,

$$BSF = 0.5 \left( \frac{TP}{a} + \frac{TN}{i} - \frac{FP}{i} - \frac{FN}{a} \right)$$

where $a$ is the number of edges and $i$ is the number of direct independencies in the true graph:

$$i = \frac{|V|(|V|-1)}{2} - a$$

where $|V|$ is the size of the variable set $V$.

There are some important differences between these three metrics. Specifically, the SHD represents classic accuracy that measures the proportion of correct classifications amongst all classifications. For example, given a ground truth graph with 1% edges and 99% direct independencies, the SHD would judge an empty graph as being 99% accurate in relation to the true graph. The F1 score relaxes this imbalance since it conveys the balance between Precision and Recall, whereas the BSF score would consider the empty graph as being 50% accurate (i.e., a score of 0) on the basis that all direct independencies have been discovered, but none of the direct dependencies (i.e., edges) have been discovered.

### B. Case studies

SaiyanH is not intended for problems that include thousands of variables, such as those in bioinformatics. As a result, the evaluation is restricted to case studies that include up to hundreds of variables.

Six real-world BNs are used to generate synthetic data. Three of them represent traditional and widely used case studies, whereas the other three come from recent real-world BN applications. The case studies represent a mixture of simple and complex models. Specifically,

1. Asia: A small network designed for patient diagnosis [25]. It consists of eight nodes, eight arcs, 18 free parameters, and has a maximum in-degree of two.

2. Alarm: A medium network designed for an alarm message monitoring system [26]. It consists of 37 nodes, 46 arcs, 509 free parameters and has a maximum in-degree of four.

3. Pathfinder: A very large network designed for decision support in surgical pathology [27]. It consists of 109 nodes, 195 arcs, 71890 free parameters, and has a maximum in-degree of five.

4. Sports: A small real-world BN that combines a rating system with various team performance statistics to predict match scores in football [28]. It consists of nine nodes, 15 arcs, 1049 free parameters, and has a maximum in-degree of two.

5. ForMed: A large real-world BN designed for risk management of violent reoffending in mentally ill prisoners [29]. It consists of 88 nodes, 138 arcs, 912 free parameters, and has a maximum in-degree of six.

6. Property: A medium real-world BN designed for the assessment of investment decisions in the UK property market [30]. It consists of 27 nodes, 31 arcs, 3056 free parameters, and has a maximum in-degree of three.

### B. Structure learning algorithms considered

The learning performance of SaiyanH is assessed with reference to other 12 algorithms that have been applied to the same data. The algorithms selected represent state-of-the-art or well-established implementations that have also been tested in a larger relevant study [10]. Specifically,

1. PC-Stable: the modern stable version of the most popular constraint-based algorithm called PC that resolves the issue on the order dependency of the variables in the data [31] [32].

2. FCI: which is PC extended to account the possibility of latent variables in the data [33].

3. FGES: an efficient version of the popular score-based GES algorithm that was developed by Meek [34] and further improved by Chickering [35].

4. GFCI: a hybrid learning algorithm that combines the FCI and FGES algorithms [36].

5. RFCI-BSC: a hybrid version of the constraint-based RFCI that improves accuracy via model averaging [20]. This is a non-deterministic algorithm that produces a slightly different result each time it is executed. The results of RFCI-BSC represent the average score across 10 executions, for each experiment.

6. Inter-IAMB: an improved version of IAMB that avoids false positives in the Markov Blanket detection phase [18].

7. MMHC: perhaps the most popular hybrid learning algorithm [24]. It combines the constraint-based MMPC with hill-climbing search.

8. GS: a constraint-based algorithm that recovers the Markov blanket of each node based on pairwise independence test [37].

9. HC: a score-based hill-climbing search algorithm that tends to terminate in a local maximum [38].

10. TABU: a score-based algorithm that extends HC with Tabu search. While TABU also tends to

11. H2PC: a hybrid learning algorithm that combines the constraint-based HPC and score-based HC algorithms [19].

12. ILP: an integer linear programming score-based approach that returns the graph that maximises the global score of a scoring function [39].

The R package *r-causal* v1.1.1 which makes use of the TETRAD freeware implementation [40] was used to test algorithms 1 to 5. The *bnlearn* R statistical package version 4.5 [41] was used to test algorithms 6 to 11. Finally, ILP was tested using the GOBNILP software [42]. All algorithms have been used with their hyperparameter defaults as implemented in each software. A six-hour runtime limit is applied to each algorithm for each experiment.

## IV. RESULTS AND DISCUSSION

### A. Accuracy of the learned graphs

Fig 2 presents the accuracy scores of SaiyanH with reference to the scores produced by the other 12 algorithms. Each of the 18 graphs corresponds to a case study and a scoring metric (i.e., six case studies over three scoring metrics). The *y*-axis of each graph represents the metric score, whereas the *x*-axis represents the fives sample sizes of the input data. Note that, in contrast to F1 and BSF scores, a lower SHD score represents a better performance. Cases in which an algorithm failed to produce a result within the six-hour runtime limit are illustrated with incomplete lines in each graph.

The results suggest that all algorithms tend to improve learning accuracy with sample size and are rather consistent across all case studies. For example, it is usually the case that the best performance is found with either 100k or 1000k samples. However, the case studies differ in complexity which means that the same sample size can be large for simple networks and small for complex networks. For example, the sample size of 10k is large for Asia, which is the simplest case study with just 18 free parameters, and small for Pathfinder which is the most complex case study with 71890 free parameters. This explains why in the case of Asia the performance of the algorithms maximises once the sample size of the input data reaches 10k observations. Conversely, the performance of the algorithms continues to improve with the sample size in the case of Pathfinder. Moreover, all algorithms show considerably worse performance on the Pathfinder case study compared to all the other case studies.

TABLE 1
THE AVERAGE RANK ACHIEVED BY EACH OF THE ALGORITHMS OVER ALL CASE STUDIES AND ACROSS ALL SAMPLE SIZES OF THE INPUT DATA.

| Rank | Algorithm | F1 Rank | SHD Rank | BSF Rank | Overall rank |
|---|---|---|---|---|---|
| 1 | TABU | 3.07 | 4.20 | 2.90 | 3.39 |
| 2 | HC | 3.43 | 4.53 | 3.03 | 3.67 |
| 3 | ILP | 4.40 | 6.03 | 3.73 | 4.72 |
| 4 | SaiyanH | 4.87 | 7.33 | 4.37 | 5.52 |
| 5 | H2PC | 5.60 | 4.80 | 6.37 | 5.59 |
| 6 | GFCI | 6.17 | 6.50 | 6.37 | 6.34 |
| 7 | FCI | 7.07 | 6.17 | 7.03 | 6.76 |
| 8 | FGES | 6.80 | 7.33 | 6.50 | 6.88 |
| 9 | MMHC | 7.07 | 6.07 | 7.83 | 6.99 |
| 10 | PC-Stable | 7.43 | 6.37 | 7.33 | 7.04 |
| 11 | Int-IAMB | 9.20 | 8.10 | 9.53 | 8.94 |
| 12 | RFCI-BSC | 10.87 | 8.33 | 10.77 | 9.99 |
| 13 | GS | 10.83 | 9.67 | 10.80 | 10.43 |

In contrast, some of the scoring metrics provide conflicting conclusions about the relative accuracy between algorithms. For example, the SHD metric occasionally ranks SaiyanH well below average when the sample size of the input data is lowest, and these results contradict the F1 and BSF metrics which rank SaiyanH well above average for the same experiments. The contradiction between these metrics extends to many other algorithms. This phenomenon arises because the SHD metric represents classification accuracy which tends to be biased in favour of graphs which incorporate a limited number of edges [10, 23].

Table 1 summarises the performance of the algorithms over each case study and across all metrics. Consistent with the above discussion, the results show that SaiyanH performed very good in terms of F1 and BSF scores, and below average in terms of SHD score. Overall, SaiyanH ranked 4[th] and outperformed algorithms such as FGES, MMHC and PC which tend to be used for benchmarking new algorithms in this field of research. Interestingly, the performance of the top three algorithms is fully driven by score-based learning.

### B. Analysis of the edges and independent subgraphs

Table 2 presents the number of independent subgraphs generated by each of the algorithms for each case study and sample size. As intended, SaiyanH generates a single connected DAG in all the experiments. On the other hand, the other algorithms routinely generate multiple subgraphs despite all the input variables being dependent in all case studies. This observation extends to very simple networks. For example, while the Asia network consists of just 18 free parameters, none of the other algorithms managed to produce a connected graph when the sample size was lowest, and only five of the other algorithms returned a connected graph when the sample size was highest.

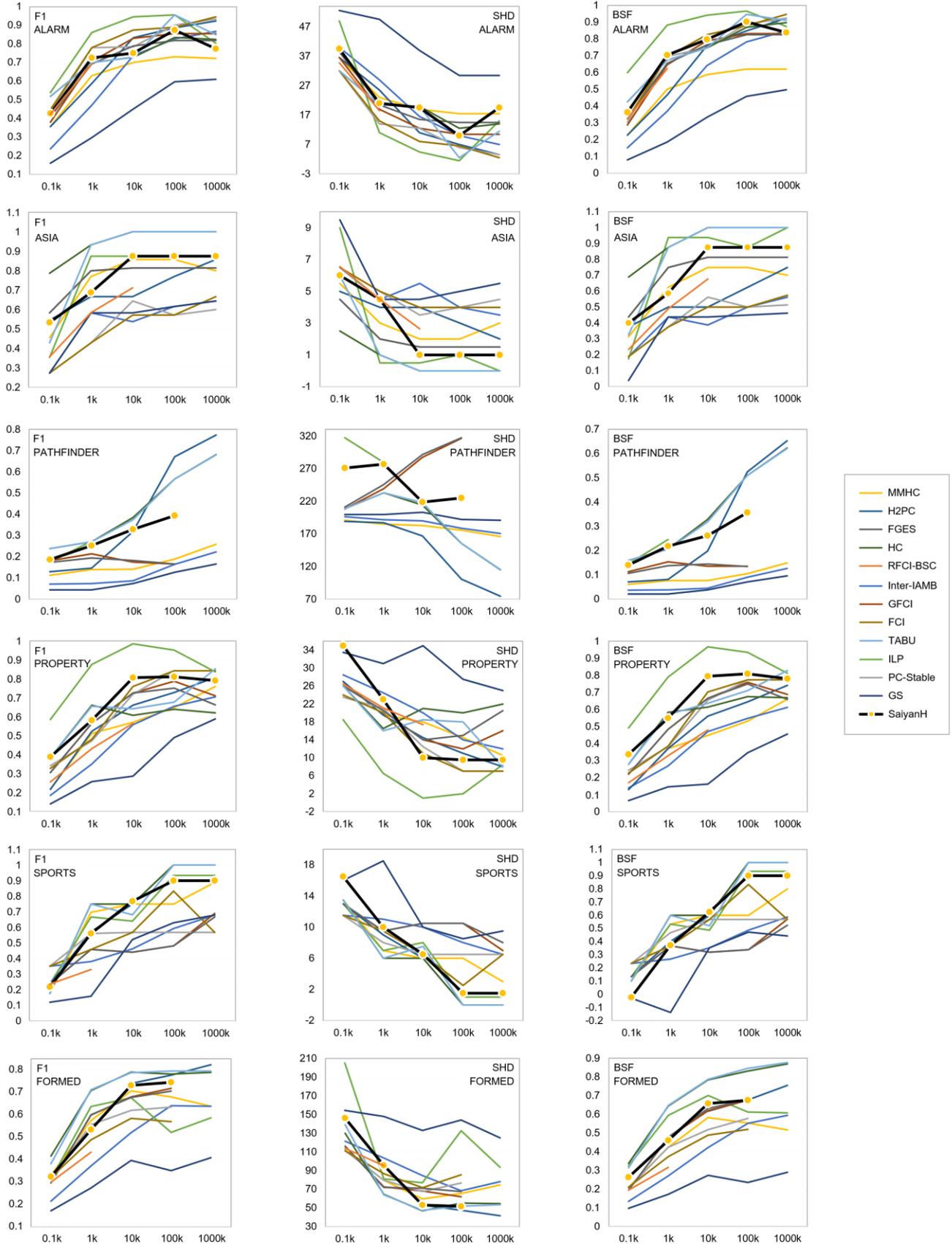

**FIGURE 2.** Performance of SaiyanH given F1 (harmonic mean of Recall and Precision), SHD, and BSF scores, over six case studies, five sample sizes (0.1k to 1000k samples) per case study, and with reference to the performance of the other 12 algorithms.

TABLE 2
THE NUMBER OF INDEPENDENT SUBGRAPHS GENERATED BY EACH ALGORITHM FOR EACH CASE STUDY AND SAMPLE SIZE. N/A INDICATES THAT THE ALGORITHM DID NOT COMPLETE LEARNING WITHIN THE SIX-HOUR LIMIT.

| Algorithm | Case study | 0.1k | 1k | 10k | 100k | 1000k |
|---|---|---|---|---|---|---|
| SaiyanH | Alarm | 1 | 1 | 1 | 1 | 1 |
| ILP | Alarm | 1 | 2 | 2 | 2 | 1 |
| FGES | Alarm | 10 | 3 | 2 | 2 | 2 |
| Inter-IAMB | Alarm | 24 | 11 | 7 | 3 | 1 |
| H2PC | Alarm | 24 | 10 | 3 | 2 | 1 |
| PC-Stable | Alarm | 18 | 5 | 3 | 2 | 2 |
| FCI | Alarm | 16 | 5 | 3 | 2 | 1 |
| GFCI | Alarm | 12 | 3 | 2 | 2 | 2 |
| RFCI-BSC | Alarm | 13.3 | 5.2 | n/a | n/a | n/a |
| MMHC | Alarm | 24 | 10 | 8 | 7 | 7 |
| GS | Alarm | 27 | 19 | 13 | 12 | 9 |
| HC | Alarm | 7 | 3 | 3 | 2 | 1 |
| TABU | Alarm | 7 | 3 | 3 | 2 | 1 |
| SaiyanH | Asia | 1 | 1 | 1 | 1 | 1 |
| ILP | Asia | 2 | 1 | 1 | 1 | 1 |
| FGES | Asia | 4 | 2 | 1 | 1 | 1 |
| Inter-IAMB | Asia | 5 | 4 | 3 | 3 | 2 |
| H2PC | Asia | 5 | 4 | 4 | 3 | 2 |
| PC-Stable | Asia | 5 | 2 | 2 | 2 | 2 |
| FCI | Asia | 5 | 2 | 2 | 2 | 2 |
| GFCI | Asia | 4 | 2 | 1 | 1 | 1 |
| RFCI-BSC | Asia | 4 | 2 | 1.2 | n/a | n/a |
| MMHC | Asia | 5 | 3 | 2 | 2 | 2 |
| GS | Asia | 5 | 4 | 4 | 3 | 2 |
| HC | Asia | 2 | 2 | 1 | 1 | 1 |
| TABU | Asia | 2 | 2 | 1 | 1 | 1 |
| SaiyanH | Pathfinder | 1 | 1 | 1 | 1 | n/a |
| ILP | Pathfinder | 19 | 6 | n/a | n/a | n/a |
| FGES | Pathfinder | 42 | 16 | 5 | 5 | n/a |
| Inter-IAMB | Pathfinder | 85 | 90 | 87 | 81 | 72 |
| H2PC | Pathfinder | 87 | 85 | 62 | 17 | 8 |
| PC-Stable | Pathfinder | n/a | n/a | n/a | n/a | n/a |
| FCI | Pathfinder | 80 | n/a | n/a | n/a | n/a |
| GFCI | Pathfinder | 42 | 16 | 5 | 5 | n/a |
| RFCI-BSC | Pathfinder | 70.3 | n/a | n/a | n/a | n/a |
| MMHC | Pathfinder | 89 | 89 | 91 | 87 | 81 |
| GS | Pathfinder | 100 | 100 | 92 | 85 | 78 |
| HC | Pathfinder | 33 | 13 | 3 | 2 | 2 |
| TABU | Pathfinder | 33 | 13 | 3 | 2 | 2 |
| SaiyanH | Property | 1 | 1 | 1 | 1 | 1 |
| ILP | Property | 8 | 5 | 1 | 1 | 1 |
| FGES | Property | 12 | 5 | 5 | 1 | 1 |
| Inter-IAMB | Property | 20 | 15 | 9 | 7 | 6 |
| H2PC | Property | 21 | 14 | 7 | 6 | 4 |
| PC-Stable | Property | 14 | 8 | 4 | 4 | 4 |
| FCI | Property | 15 | 9 | 3 | 4 | 4 |
| GFCI | Property | 12 | 5 | 5 | 1 | 1 |
| RFCI-BSC | Property | 16.4 | 11.6 | 6.8 | n/a | n/a |
| MMHC | Property | 21 | 13 | 9 | 8 | 6 |
| GS | Property | 22 | 20 | 17 | 12 | 7 |
| HC | Property | 11 | 5 | 2 | 1 | 1 |
| TABU | Property | 11 | 5 | 2 | 1 | 1 |
| SaiyanH | Sports | 1 | 1 | 1 | 1 | 1 |
| ILP | Sports | 7 | 1 | 1 | 1 | 1 |
| FGES | Sports | 7 | 1 | 1 | 1 | 1 |
| Inter-IAMB | Sports | 4 | 3 | 1 | 1 | 1 |
| H2PC | Sports | 7 | 3 | 1 | 1 | 1 |
| PC-Stable | Sports | 4 | 1 | 1 | 1 | 1 |
| FCI | Sports | 4 | 1 | 1 | 1 | 1 |
| GFCI | Sports | 7 | 1 | 1 | 1 | 1 |
| RFCI-BSC | Sports | n/a | n/a | n/a | n/a | n/a |
| MMHC | Sports | 7 | 1 | 1 | 1 | 1 |
| GS | Sports | 7 | 5 | 1 | 1 | 1 |
| HC | Sports | 7 | 1 | 1 | 1 | 1 |
| TABU | Sports | 7 | 1 | 1 | 1 | 1 |
| SaiyanH | ForMed | 1 | 1 | 1 | 1 | n/a |
| ILP | ForMed | 6 | 4 | 1 | 1 | 1 |
| FGES | ForMed | 43 | 13 | 2 | 2 | n/a |
| Inter-IAMB | ForMed | 53 | 25 | 13 | 9 | 5 |
| H2PC | ForMed | n/a | n/a | 9 | 5 | 4 |
| PC-Stable | ForMed | 47 | 19 | 10 | 6 | n/a |
| FCI | ForMed | 45 | 19 | 10 | 6 | n/a |
| GFCI | ForMed | 43 | 13 | 2 | 2 | n/a |
| RFCI-BSC | ForMed | 46.3 | 26.4 | n/a | n/a | n/a |
| MMHC | ForMed | 50 | 25 | 12 | 16 | 16 |
| GS | ForMed | 57 | 43 | 31 | 36 | 31 |
| HC | ForMed | 8 | 4 | 1 | 1 | 1 |
| TABU | ForMed | 7 | 4 | 1 | 1 | 1 |

Moreover, the number of independent subgraphs produced by some of the other algorithms increases substantially with the complexity of the true graph. The most extreme example involves the GS algorithm when applied to the Pathfinder case, where it produced 100 subgraphs for the lowest sample size and 68 subgraphs for the highest sample size. Remarkably, and further to what has been discussed in subsection IV.A, the outcome of 100 subgraphs was ranked highly by the SHD metric. In contrast, the F1 and BSF metrics ranked this outcome lowest.

While most of the other algorithms generate several subgraphs in most of the experiments, TABU and HC did well since in many cases they had correctly identified that the input variables are dependent. This also partly explains why the TABU and HC algorithms outperformed all the other algorithms in results of Table 1, and this is an interesting outcome considering that most of the other score-based and hybrid learning algorithms (including SaiyanH) already use some form of HC search to explore the search space of graphs.

Fig 3 analyses the number of edges produced by each of the algorithms and their relation to the number of edges in the true graphs. Each graph in Fig 3 corresponds to a case study. Each case study and algorithm associate with a range Δ, where Δ is the difference between the number of edges learned and the number of edges in the true graph, and the interval Δ represents the minimum and the maximum discrepancy across all the five experiments in each case study (i.e., over all the five sample sizes). Note that failed attempts by an algorithm to produce a graph are excluded. When this happens, the number of actual experiments on which the interval is based is superimposed above the interval. For example, in the Asia case study, the RFCI-BSC algorithm failed to produce a result in two out of the five experiments and thus, its interval Δ indicates that it was based on just three experiments. It is important to clarify that the failed attempts of an algorithm always occur for the highest sample sizes. As a result, when an interval Δ is based on less than five experiments, it tends to underestimates the number of learned edges since higher sample sizes tend to produce more edges.

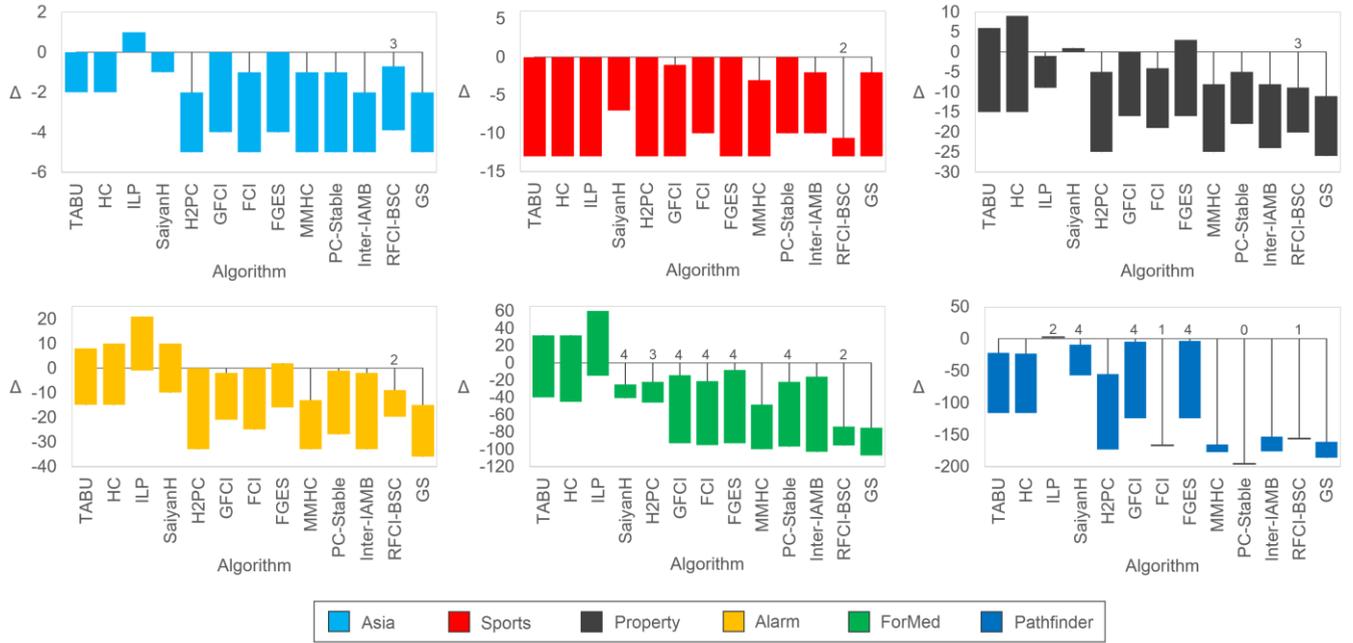

**FIGURE 3.** The number of learned edges with respect to the number of true edges. The results are presented for each algorithm and over all five sample sizes per case study, where Δ is the discrepancy between learned and true edges. Failed attempts by the algorithms to produce a graph are excluded. Intervals Δ based on less than five experiments indicate the number of actual experiments above the interval.

The edge analysis in Fig 3 provides insights into potential underfitting and overfitting issues. In fact, the results suggest that most algorithms do underfit the graphs, at least in terms of the number of edges produced. These results are consistent with the number of independent subgraphs depicted in Table 2. While the risk of underfitting increases with fewer samples in the input data, the results suggest that underfitting persists across all sample sizes tested. For example, while the Asia network consists of just 18 free parameters, most of the algorithms reveal an underfitting trend across all sample sizes. Likewise, underfitting appears to increase in severity with the complexity of the network. For example, the Inter-IAMB and GS algorithms discovered a maximum of 42 and 34 edges respectively (at 1000k samples), out of the 195 true edges in the Pathfinder case study. Conversely, no algorithm overfitted the graphs. One algorithm that did show some tendency towards overfitting, however, is ILP. This happened on the Alarm and ForMed case studies. However, ILP did very well in minimising the discrepancy Δ across all the six case studies.

Starting from the case study with the least number of edges, the algorithms that performed best in terms of minimising discrepancy Δ, as well as the interval Δ across the different sample sizes in each case study, are:

1. Asia: the ILP and SaiyanH algorithms (with TABU and HC closely behind) with a minimum discrepancy Δ of 0, a maximum discrepancy Δ of ±1, and a range Δ of 1 for both algorithms (from +1 to 0 and from 0 to -1 respectively).

2. Sports: the SaiyanH algorithm with a minimum discrepancy Δ of 0, a maximum discrepancy Δ of -7 and a range Δ of 7.

3. Property: the SaiyanH algorithm with a minimum discrepancy Δ of 0, a maximum discrepancy Δ of +1, and a range Δ of 1.

4. Alarm: the SaiyanH algorithm (with TABU and HC closely behind) with a minimum discrepancy Δ of -10, a maximum discrepancy Δ of 10, and a range Δ of 20.

5. ForMed: the TABU (with HC closely behind) with a minimum discrepancy Δ of -40, a maximum discrepancy Δ of 32, and a range Δ of 72.

6. Pathfinder: ILP with a minimum Δ of 0, a maximum Δ of 3, and a range Δ of 3. However, note that ILP's result in Pathfinder is based on just two experiments.

Overall, the edge statistics are in agreement with the results in Table 1, in that the top four algorithms are also the ones that best approximate the number of true edges. The restriction in SaiyanH to produce a connected DAG has helped the algorithm to avoid underfitting, as well as to perform best in terms of minimising discrepancy Δ.

## C. Time complexity

The time complexity of SaiyanH is provided in Table 3. The results show that its runtime increases rapidly with the number of nodes and the sample size of the input data. SaiyanH failed to produce a graph within the six-hour time limit in two out of the 30 experiments. This was also the case for some of the other algorithms (refer to Fig 2). It is worth noting that the Pathfinder case study (indicated with 109 nodes in Table 3) includes a variable with 63 states, which is rather unusual for discrete BNs and can influence time complexity in different ways depending on the learning process of the algorithm.

Table 4 extends the information on time complexity by presenting the proportion of time SaiyanH spent to complete each of the three learning phases, for each case study and sample size combination. This information is useful for two reasons. Firstly, it highlights which parts of this new implementation may be inefficient. Secondly, it reveals how the relative proportion of runtime varies between the different learning phases given the number of variables in conjunction the sample size of the input data.

The results from the Pathfinder and ForMed case studies, which required the most runtime, suggest that the constraint-based learning of phase 2 is responsible for 77% to 94% of the total runtime. This outcome suggests that the constraint-based phase in SaiyanH does not scale well with the number of variables and the sample size of the input data. One reason why constraint-based learning is inefficient in SaiyanH is because conditional independence tests are performed over all possible triples, including testing for both $V_A \rightarrow V_B | V_C$ and $V_B \rightarrow V_A | V_C$ as defined by the MMD score. Therefore, the efficiency of SaiyanH could be improved via pruning of conditional independence tests. However, the effectiveness of this type of pruning is hard to predict, both in terms of possible gains in speed and the impact on the accuracy of the learned graph. This is because the conditional independence classifications from phase 2 are also used to prune the search space of DAGs in phase 3. For example, Table 4 shows that when the sample size of the input data is low, constraint-based learning only accounts for up to a third of the total runtime whereas score-based learning becomes the most time-consuming phase of the algorithm.

TABLE 3
TIME COMPLEXITY OF SAIYANH BASED ON A SINGLE-CORE (TURBO BOOST) SPEED OF 4.7GHZ.

| Nodes | True edges | Max in-degree | # free param | Sample size | Runtime (sec) |
|---|---|---|---|---|---|
| 37 | 46 | 4 | 509 | 0.1k | 1 |
| 8 | 8 | 2 | 18 | 0.1k | 1 |
| 8 | 8 | 2 | 18 | 1k | 1 |
| 8 | 8 | 2 | 18 | 10k | 1 |
| 8 | 8 | 2 | 18 | 100k | 1 |
| 27 | 31 | 3 | 3,056 | 1k | 1 |
| 9 | 15 | 2 | 1,049 | 0.1k | 1 |
| 9 | 15 | 2 | 1,049 | 1k | 1 |
| 9 | 15 | 2 | 1,049 | 10k | 1 |
| 37 | 46 | 4 | 509 | 1k | 5 |
| 27 | 31 | 3 | 3,056 | 10k | 9 |
| 27 | 31 | 3 | 3,056 | 0.1k | 10 |
| 9 | 15 | 2 | 1,049 | 100k | 10 |
| 8 | 8 | 2 | 18 | 1m | 13 |
| 37 | 46 | 4 | 509 | 10k | 16 |
| 109 | 195 | 5 | 71,890 | 0.1k | 30 |
| 88 | 138 | 6 | 912 | 0.1k | 48 |
| 88 | 138 | 6 | 912 | 1k | 49 |
| 27 | 31 | 3 | 3,056 | 100k | 101 |
| 9 | 15 | 2 | 1,049 | 1000k | 120 |
| 109 | 195 | 5 | 71,890 | 1k | 135 |
| 37 | 46 | 4 | 509 | 100k | 213 |
| 88 | 138 | 6 | 912 | 10k | 220 |
| 109 | 195 | 5 | 71,890 | 10k | 521 |
| 27 | 31 | 3 | 3,056 | 1000k | 1,400 |
| 37 | 46 | 4 | 509 | 1000k | 2,900 |
| 88 | 138 | 6 | 912 | 100k | 3,494 |
| 109 | 195 | 5 | 71,890 | 100k | 12,043 |
| 109 | 195 | 5 | 71,890 | 1000k | >6h |
| 88 | 138 | 6 | 912 | 1000k | >6h |

TABLE 4
TIME COMPLEXITY OF SAIYANH IN TERMS OF THE PERCENTAGE OF TIME SPENT TO COMPLETE EACH OF THE THREE LEARNING PHASES. CASES WITH RUNTIME UP TO 1SEC ARE EXCLUDED. N/A INDICATES THAT THE ALGORITHM DID NOT COMPLETE LEARNING WITHIN THE SIX-HOUR LIMIT.

| Case (phase) | 0.1k | 1k | 10k | 100k | 1000k |
|---|---|---|---|---|---|
| Alarm (1) | - | 0% | 6% | 13% | 9% |
| Asia (1) | - | - | - | - | 31% |
| Pathfinder (1) | 0% | 1% | 3% | 3% | n/a |
| Property (1) | 0% | - | 0% | 19% | 15% |
| Sports (1) | - | - | - | 10% | 13% |
| ForMed (1) | 0% | 0% | 3% | 5% | n/a |
| Alarm (2) | - | 20% | 69% | 78% | 81% |
| Asia (2) | - | - | - | - | 69% |
| Pathfinder (2) | 17% | 28% | 82% | 77% | n/a |
| Property (2) | 0% | - | 89% | 76% | 77% |
| Sports (2) | - | - | - | 30% | 23% |
| ForMed (2) | 6% | 37% | 89% | 94% | n/a |
| Alarm (3) | - | 80% | 25% | 9% | 10% |
| Asia (3) | - | - | - | - | 0% |
| Pathfinder (3) | 83% | 71% | 15% | 21% | n/a |
| Property (3) | 100% | - | 11% | 5% | 8% |
| Sports (3) | - | - | - | 60% | 65% |
| ForMed (3) | 94% | 63% | 8% | 1% | n/a |

## V. CONCLUDING REMARKS AND FUTURE WORK

This paper described a novel hybrid BN structure learning algorithm that relies on restrictions in the search space of DAGs to produce a graph that enables full propagation of evidence. The learning restriction is imposed under the controversial assumption that the data variables are dependent. The implementation of the algorithm [43] as well as the datasets used in this study are available online at www.bayesys.com.

Clearly, this algorithm is unsuitable in problems where we seek to discover whether the input variables are dependent. However, it becomes useful in real-world problems where decision makers desire a model that enables full propagation of evidence. The empirical results show that almost all the other algorithms would never connect all the variables of the input data when the sample size of the data is low relative the dimensionality of the model, despite these variables being dependent in the true graph. This is a problem because real data are often limited in terms of sample size and rich in terms of the number of the variables (i.e., in dimensionality). Therefore, the benefit of assuming that the input variables are dependent comes in the form of practical usefulness that can be viewed as a knowledge-based constraint.

Because SaiyanH is a novel implementation, it comes with both limitations and potential for improvement. Firstly, its application is limited to discrete and complete datasets. Moreover, while SaiyanH performed best in estimating the number of true edges, and well in terms of F1 and BSF scores, it did not do so well in terms of SHD score and particularly when the sample size of the input data was lowest. This observation suggests that some of the forced edges generated to ensure the DAG output is connected, are not correct at the same rate as those generated unrestrictedly.

Lastly, the results presented in this paper are based on unoptimised cut-off dependency thresholds adopted by other constraint-based algorithms (refer to Section II.B). This ensured that the comparison between algorithms is as fair as possible, since all algorithms have been examined with their hyperparameter defaults. However, because SaiyanH is based on an unconventional dependency function, it may benefit from cut-off thresholds that differ to those used as hyperparameter defaults in other algorithms. In investigating the value of constraint-based relative to its time complexity, future work will also explore the impact of parameter optimisation on conditional independence classifications.

## ACKNOWLEDGEMENTS

This research was supported by the ERSRC Fellowship project EP/S001646/1 on Bayesian Artificial Intelligence for Decision Making under Uncertainty, and by The Alan Turing Institute in the UK under the EPSRC grant EP/N510129/1.

## REFERENCES

[1] J. Pearl. "Reverend Bayes on Inference Engines: A Distributed Hierarchical Approach". In *Proceedings of the 2nd AAAI Conference on Artificial Intelligence*, pages 133–136, Pittsburgh, Pennsylvania, August 1982. AAAI Press.

[2] J. Pearl. "Bayesian Networks: A model of self-activated memory for evidential reasoning". In *Proceedings of the 7th Conference of the Cognitive Science Society*, pp. 329–334, 1985.

[3] D. M. Chickering, D. Heckerman and C. Meek. "Large-sample learning of Bayesian networks is NP-hard". *Journal of Machine Learning Research*, vol. 5, pp. 1287–1330, 2004.

[4] P. Spirtes, C. Glymour and R. Scheines. "*Causation, Prediction, and Search*". New York: Springer-Verlag, 1993.

[5] P. Spirtes, C. Glymour and R. Scheines. "Reply to Humphreys and Freedman's Review of Causation, Prediction, and Search." *British Journal of the Philosophy of Science*, vol. 48, pp. 555–568, 1997.

[6] P. Humphreys and D. Freedman. "The Grand Leap". *British Journal of the Philosophy of Science*, vol. 47, pp. 113–123, 1996.

[7] K. B. Korb and C. S. Wallace. "In Search of the Philosopher's Stone: Remarks on Humphreys and Freedman's Critique of Causal Discovery". *British Journal of the Philosophy of Science*, vol. 48, pp. 543–553, 1997.

[8] D. Freedman and P. Humphreys. "Are there algorithms that discover causal structure". *Synthese*, vol. 121, pp. 29–54, 1999.

[9] K. Korb, and A. Nicholson. "*Bayesian Artificial Intelligence (Second Edition)*". CRC Press, London, UK, 2011.

[10] A. C. Constantinou, Y. Liu, K. Chobtham, Z. Guo and N. K. Kitson. "Large-scale empirical validation of Bayesian Network structure learning algorithms with noisy data". *arXiv:2005.09020* [cs.LG], 2020.

[11] D. Heckerman, D. Geiger, and D. M. Chickering. "Learning Bayesian networks: The combination of knowledge and statistical data". *Machine Learning*, vol. 20, pp. 197–243, 1995.

[12] N. Fenton, M. Neil, D. Lagnado, W. Marsh, B. Yet, and A. Constantinou. How to model mutually exclusive events based on independent causal pathways in Bayesian network models. *Knowledge-Based Systems*, vol. 133, pp. 39–50, 2016.

[13] A. C. Li, and P. van Beek. "Bayesian Network Structure Learning with Side Constraints". In *Proceedings of Machine Learning Research*, vol. 72, pp. 225–236, 2018

[14] L. M. de Campos and J. G. Castellano. "Bayesian network learning algorithms using structural restrictions". *International Journal of Approximate Reasoning*, vol. 45, pp. 233–254, 2007.

[15] M. J. Flores, A. E. Nicholson, A. Brunskill, K. B. Korb, and S. Mascaro. "Incorporating expert knowledge when learning Bayesian network structure: a medical case study". *Artificial intelligence in medicine*, vol. 53, Iss. 3, pp. 181–204, 2011.

[16] A. Constantinou. "Learning Bayesian Networks with the Saiyan algorithm". *ACM Transactions on Knowledge Discovery from Data*, 2020.

[17] A. Darwiche. "*Modeling and reasoning with Bayesian Networks*". Cambridge University Press, 2009.

[18] S. Yaramakala and D. Margaritis. "Speculative Markov Blanket Discovery for Optimal Feature Selection". In *Proceedings of the Fifth IEEE International Conference on Data Mining*, pp. 809–812, 2005.

[19] M. Gasse, A. Aussem and H. Elghazel. "A Hybrid Algorithm for Bayesian Network Structure Learning with Application to Multi-Label Learning". *Expert Systems with Applications*, vol. 41, Iss. 15, pp. 6755–6772, 2014.

[20] F. Jabbari, J. Ramsey, P. Spirtes and G. Cooper. "Discovery of causal models that contain latent variables through Bayesian scoring of independence constraints". *Machine Learning and Knowledge Discovery in Databases*, pp. 142–157, 2017.

[21] K. Chobtham and A. Constantinou. "Bayesian network structure learning with causal effects in the presence of latent variables". *arXiv:2005.14381* [cs.LG], 2020.

[22] J. Pearl. "*Causality*". Cambridge University Press, 2009.

[23] A. C. Constantinou. "Evaluating structure learning algorithms with a balanced scoring function". *arXiv 1905.12666* [cs.LG], 2019.

[24] I. Tsamardinos, L. E. Brown and C. F. Aliferis. "The Max-Min Hill-Climbing Bayesian Network Structure Learning Algorithm". *Machine Learning,* vol. 65, pp. 31–78, 2006.

[25] S. Lauritzen and D. Spiegelhalter. "Local Computation with Probabilities on Graphical Structures and their Application to Expert Systems (with discussion)". *Journal of the Royal Statistical Society: Series B (Statistical Methodology)*, vol. 50, pp. 157–224, 1998.

[26] I. A. Beinlich, H. J. Suermondt, R. M. Chavez and G. F. Cooper. "The ALARM Monitoring System: A Case Study with Two Probabilistic Inference Techniques for Belief Networks". In *Proceedings of the 2nd European Conference on Artificial Intelligence in Medicine*, pp. 247–256, 1989.

[27] D. Heckerman, E. Horwitz and B. Nathwani. "Towards Normative Expert Systems: Part I. The Pathfinder Project". *Methods of Information in Medicine*, vol. 31, pp. 90–105, 1992.

[28] A. Constantinou. "Asian handicap football betting with rating-based hybrid Bayesian networks". *arXiv:2003.09384* [stat.AP], 2019.

[29] A. C. Constantinou, M. Freestone, W. Marsh, N. Fenton, and J. Coid. "Risk assessment and risk management of violent reoffending among prisoners". *Expert Systems with Applications*, vol. 42, Iss. 21, pp. 7511–7529, 2015.

[30] A. C. Constantinou and N. Fenton. "The future of the London Buy-To-Let property market: Simulation with Temporal Bayesian Networks". *PloS ONE*, 12(6): e0179297, 2017.

[31] P. Spirtes and C. Glymour. "An algorithm for fast recovery of sparse causal graphs". *Social Science Computer Review*, vol. 9, Iss. 1, 1991.

[32] D. Colombo and M. H. Maathuis. "Order-Independent Constraint-Based Causal Structure Learning". *Journal of Machine Learning Research*, vol. 15, pp. 3921–3962, 2014.

[33] P. Spirtes, C. Meek, and T. Richardson. "An algorithm for causal inference in the presence of latent variables and selection bias". In Clark Glymour and Gregory Cooper (Eds.), *Computation, Causation, and Discovery*. The MIT Press, Cambridge, MA, pp. 211–252, 1999.

[34] C. Meek. "Graphical Models: Selecting causal and statistical models". PhD dissertation, *Carnegie Mellon University*, 1997.

[35] D. M. Chickering. "Optimal structure identification with greedy search". *Journal of Machine Learning Research,* vol. 3, pp. 507–554, 2002.

[36] J. M. Ogarrio, P. Spirtes and J. Ramsey. "A Hybrid Causal Search Algorithm for Latent Variable Models". In *Proceedings of the Eighth International Conference on Probabilistic Graphical Models*, vol. 52, pp. 368–379, 2016.

[37] D. Margaritis. "Learning Bayesian Network Model Structure from Data". PhD dissertation, *School of Computer Science, Carnegie-Mellon University*, Pittsburgh, PA, 2003.

[38] M. Scutari, C. Vitolo, A. Tucker, A. "Learning Bayesian networks from big data with greedy search: computational complexity and efficient implementation". *Statistics and Computing*, vol. 29, pp 1095–1108, 2019.

[39] J. Cussens. "Bayesian network learning with cutting planes". In *Proceedings of the Twenty-Seventh Conference on Uncertainty in Artificial Intelligence*, pp. 153–160, July 2011.

[40] C. Wongchokprasitti. "*R-causal R Wrapper for Tetrad Library*", v1.1.1, 2016. [Online] Available https://github.com/bd2kccd/r-causal

[41] M. Scutari and R. Ness. "*Package 'bnlearn'*". *CRAN*, 2019.

[42] J. Cussens. "*GOBNILP 1.6.2 User/Developer Manual*", University of York, UK, 2015. [Online] Available: https://www.cs.york.ac.uk/aig/sw/gobnilp/manual.pdf

[43] A. Constantinou. "*The Bayesys user manual*". Queen Mary University of London, London, 2020. [Online] Available: http://bayesian-ai.eecs.qmul.ac.uk/bayesys/

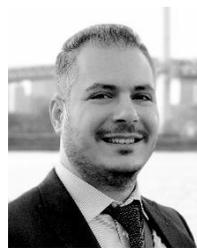

**ANTHONY C. CONSTANTINOU** is a Senior Lecturer at Queen Mary University of London, where he leads the Bayesian Artificial Intelligence research lab. He holds a Fellowship from the EPSRC and a Fellowship from The Alan Turing Institute in the UK.

His research interests are in Bayesian Artificial Intelligence for causal discovery and intelligent decision making under uncertainty. He collaborates with industry partners and applies his research to a wide range of areas including sports, medicine, finance and gaming.